\documentclass[pdflatex,sn-basic,Numbered]{sn-jnl}

\usepackage{graphicx}
\usepackage{amsmath,amssymb,amsfonts}
\usepackage{booktabs}
\usepackage{multirow}
\usepackage{xcolor}
\usepackage{textcomp}
\usepackage{xspace}

\makeatletter
\def\Authorfont{\reset@font\fontsize{10bp}{12bp}\selectfont\boldmath\titraggedcenter}
\def\addressfont{\reset@font\fontsize{9bp}{11bp}\selectfont\titraggedcenter}
\makeatother

\makeatletter
\newcommand{\runinheadings}{%
  \renewcommand\subsubsection{\@startsection{subsubsection}{3}{\z@}%
    {-12pt \@plus -4pt \@minus -2pt}{-0.6em}{\normalsize\bfseries\boldmath}}%
  \renewcommand\paragraph{\@startsection{paragraph}{4}{\z@}%
    {-12pt \@plus -4pt \@minus -2pt}{-0.6em}{\normalsize\bfseries}}}
\newcommand{\blockheadings}{%
  \renewcommand\subsubsection{\@startsection{subsubsection}{3}{\z@}%
    {-12pt \@plus -4pt \@minus -2pt}{6pt}{\subsubsectionfont}}%
  \renewcommand\paragraph{\@startsection{paragraph}{4}{\z@}%
    {-12pt \@plus -4pt \@minus-2pt}{3pt}{\paragraphfont}}}
\makeatother

\makeatletter
\g@addto@macro\normalsize{%
  \abovedisplayskip 6\p@ \@plus2\p@ \@minus1\p@
  \belowdisplayskip 4\p@ \@plus2\p@ \@minus1\p@
  \abovedisplayshortskip \z@ \@plus2\p@
  \belowdisplayshortskip 4\p@ \@plus2\p@ \@minus1\p@}
\normalsize
\makeatother

\newcommand{\cholecap}{\textit{Chole-Cap-DQA}\xspace}
\newcommand{\lrygbcap}{\textit{LRYGB-Cap-DQA}\xspace}
\newcommand{\choleskill}{\textit{Chole-Skill-DQA}\xspace}
\newcommand{\lrygbskill}{\textit{LRYGB-Skill-DQA}\xspace}
\newcommand{\sd}[1]{$_{\scriptscriptstyle\pm\text{#1}}$}

\usepackage{array}
\usepackage{placeins}
\newcolumntype{L}[1]{>{\raggedright\arraybackslash}p{#1}}
\newcommand{\ph}[1]{\texttt{\{#1\}}}  %

\begin{document}
\addtocontents{toc}{\protect\setcounter{tocdepth}{-1}}  

\title[EndoComparer]{EndoComparer: Comparative Surgical Video Question Answering with Pseudo-Siamese Visual Adaptation and Perceive-then-Compare Reasoning}

\author*[1,2]{\fnm{Shi} \sur{Li}}\email{shi.li@unistra.fr}
\author[1,2]{\fnm{Lalith} \sur{Sharan}}
\author[3,4]{\fnm{Jo\"{e}l L.} \sur{Lavanchy}}
\author[1,2]{\fnm{Kun} \sur{Yuan}}
\author[1,2]{\fnm{Lorenzo} \sur{Arboit}}
\author[1,2]{\fnm{Nabani} \sur{Banik}}
\author[1,2]{\fnm{Elif} \sur{Yazici}}
\author[2,5]{\fnm{Maria} \sur{Vannucci}}
\author[1,2,7]{\fnm{Vinkle} \sur{Srivastav}}
\author[2,6]{\fnm{Pietro} \sur{Mascagni}}
\author[1,2]{\fnm{Nicolas} \sur{Padoy}}

\affil[1]{\orgdiv{ICube, UMR7357}, \orgname{University of Strasbourg, CNRS, INSERM}, \orgaddress{\city{Strasbourg}, \postcode{67000}, \country{France}}}
\affil[2]{\orgname{Institute of Image-Guided Surgery, IHU Strasbourg}, \orgaddress{\city{Strasbourg}, \postcode{67000}, \country{France}}}
\affil[3]{\orgname{University Digestive Health Care Center -- Clarunis}, \orgaddress{\city{Basel}, \postcode{4002}, \country{Switzerland}}}
\affil[4]{\orgdiv{Department of Biomedical Engineering}, \orgname{University of Basel}, \orgaddress{\city{Allschwil}, \postcode{4123}, \country{Switzerland}}}
\affil[5]{\orgdiv{General Surgery Department}, \orgname{University of Turin}, \orgaddress{\city{Turin}, \country{Italy}}}
\affil[6]{\orgname{Fondazione Policlinico Universitario Agostino Gemelli IRCCS}, \orgaddress{\city{Rome}, \country{Italy}}}
\affil[7]{\orgdiv{Indian Institute of Technology (IIT) Madras}, \orgaddress{\city{Chennai}, \country{India}}}

\abstract{
\textbf{Purpose:} Surgical training benefits from explicit comparison between cases: in patient condition, in
technique and in skill proficiency. Surgical vision-language models (VLMs) offer a promising and flexible interface
for this purpose. Yet current VLMs typically see one video at a time, and they often fall short when asked to compare
videos.
\textbf{Methods:} We propose \textit{EndoComparer}, a VLM for comparative surgical video question answering with two
components. (a) A Pseudo-Siamese Visual LoRA (PSVL) gives each video slot its own low-rank adapter over a shared
frozen vision encoder, separating the two videos in the visual pathway at the cost of a single encoder pass.
(b) A Perceive-then-Compare chain of thought (PtC-CoT) trains the model to first state what each video shows, then
derive the answer from these statements. We further curate \textit{EndoDiffVQA}, a benchmark of nearly 11K
multiple-choice questions on video-pair over cholecystectomy and gastric bypass procedures, covering anatomy, tool, action and
skill.
\textbf{Results:} EndoComparer outperforms strong generalist and surgical VLMs on detecting and specifying difference between videos. Ablations show that two proposed components contribute complimentarily, and the model remains more robust under a video-swap control that exposes confusion between the two videos.
\textbf{Conclusion:} Providing a dedicated visual reasoning path through PSVL and PtC-CoT makes cross-case
comparison of surgical videos more robust for VLMs. EndoDiffVQA and EndoComparer provide a benchmark and a strong
baseline for this task. Project page: \url{https://camma-public.github.io/endocomparer}.
}

\keywords{Surgical video question answering, Comparative reasoning, Vision-language models}

\maketitle

\section{Introduction}\label{sec:intro}
 
Surgical assessment embeds an inherent yet implicit comparative mental model: judgements of case-wise operative
difficulty or of a surgeon's skill proficiency are made against a reference, often a protocol distilled from other
cases~\citep{birkmeyer2013skill,martin1997osats,vassiliou2005goals}. Making this comparison explicit is especially
valuable during surgical training. A trainee learns more from a reference video and an account of what differs, where and when, than from being told that a dissection was ``not adequate''. A model that can state what two videos share and how they diverge would therefore provide valuable feedback, and surgical vision-language models (VLMs) are a promising interface for it. Such comparison ranges in difficulty: at one end lie factual questions about what each video shows, which instruments are used, which anatomy is exposed, which action is performed; at the other, judgements of operative skill that integrate these observations over minutes of footage.
However, VLMs typically see one video at a
time. General video VLMs~\citep{qwen3vl2025,wang2025internvl35,openai2026gpt55,google2026gemini36} answer questions
about a single video, and surgical VLMs have followed the same path, from single-frame question
answering~\citep{yuan2024ssgvqa,he2024pitvqa} to clip- and video-level
reasoning~\citep{li2024llavasurg,wang2025surgvidlm,drago2025surgvivqa,li2026surgtemp} built on surgical
vision-language pretraining~\citep{yuan2023surgvlp,yuan2024peskavlp,perez2025surglavi}; automated skill
assessment also scores one video~\citep{kiyasseh2023vision}. Whether a VLM can answer the
comparative questions surgeons actually ask, such as ``is the cystic duct dissected to the same standard in both
videos?'', has not been tested.

Comparison does not reduce to single-video understanding applied twice, because what is asked for, the relation
between the videos, is a property of neither video alone. The model must perceive each video, keep the two sets
of observations apart and bind every fact to the video it came from. General-domain benchmarks show that this
binding is where current models break: on image
pairs~\citep{fu2024blink}, image
sets~\citep{wang2024muirbench} and video pairs~\citep{burgess2025viddiff}, frontier models fall far
below humans, a failure traced to visual features that the language model cannot assign to the right
object~\citep{campbell2024binding} and to fine detail that never reaches the
text~\citep{rahmanzadehgervi2024blind}. Multi-image instruction tuning~\citep{jiang2024mantis,li2024llavainterleave}
narrows the gap but keeps a Siamese vision encoder, one set of weights applied to every input, so the two videos
are told apart only by their position in the language model. Surgery sharpens the problem: the two videos show the
same procedure, phase and instruments, so confusing which video shows what is both likely and hard to notice, and
skill differences are small and spread over clips lasting minutes. 

This paper addresses the data, the metric and the model together. We build \textit{EndoDiffVQA}, a benchmark of
10{,}939 video-pair, four-option questions over laparoscopic cholecystectomy and Roux-en-Y gastric bypass, drawn from
scene captions~\citep{nwoye2022rendezvous,lavanchy2024multibypass} and OSATS- and GOALS-style skill
scores~\citep{sharma2025early,lavanchy2024multibypass}. A deterministic pipeline turns each pair of per-video facts
into a comparative multiple-choice question whose distractors are defined failure modes rather than random
alternatives. We further propose hierarchical metrics: \textit{detect}, \textit{specify} and \textit{strict}, that separate detecting a difference from specifying it and weigh every answer equally, so that option patterns cannot be exploited and the scores show whether a model can tell the videos apart and say how they differ.

We then propose \textit{EndoComparer}, which gives each video slot a dedicated visual and reasoning path so that
observations stay bound to the right video during comparison. Pseudo-Siamese Visual LoRA (PSVL) gives each video
slot its own low-rank adapter over a shared frozen vision encoder: the two branches share the encoder weights, as in
a Siamese network, but differ in their adapters, and both run in one packed encoder pass. Perceive-then-Compare
chain of thought (PtC-CoT)~\citep{wei2022cot} trains the model to state what each video shows before answering, so
the answer letter is conditioned on the model's own per-video statements, which are supervised by the same facts the benchmark is built from, at no extra annotation cost.

Our contributions are: (i) EndoDiffVQA, the first comparative surgical video question-answering benchmark, with
detect, specify and strict metrics; (ii) EndoComparer, with Pseudo-Siamese Visual LoRA and Perceive-then-Compare
reasoning; (iii) an evaluation of eleven generalist and surgical VLMs, zero-shot and fine-tuned. EndoComparer
reaches 54.0 Strict, 5.7 points above its fine-tuned backbone and 7.1 above the best proprietary model, with the
largest gains on skill comparison, and the ablation shows the two components are complementary. Code and
our annotations on the public datasets will be released.

\section{Methods}\label{sec:methods}

\subsection{Dataset construction}\label{sec:dataset}

EndoDiffVQA turns single-video annotations into comparative multiple-choice questions over pairs of surgical
videos. It draws on two kinds of annotation, clip-level captions and expert skill scores, over two procedures,
laparoscopic cholecystectomy (Chole) and laparoscopic Roux-en-Y gastric bypass (LRYGB), giving four subsets: \cholecap, \lrygbcap, \choleskill and \lrygbskill.

\begin{figure}[t]
\centering
\includegraphics[width=\textwidth]{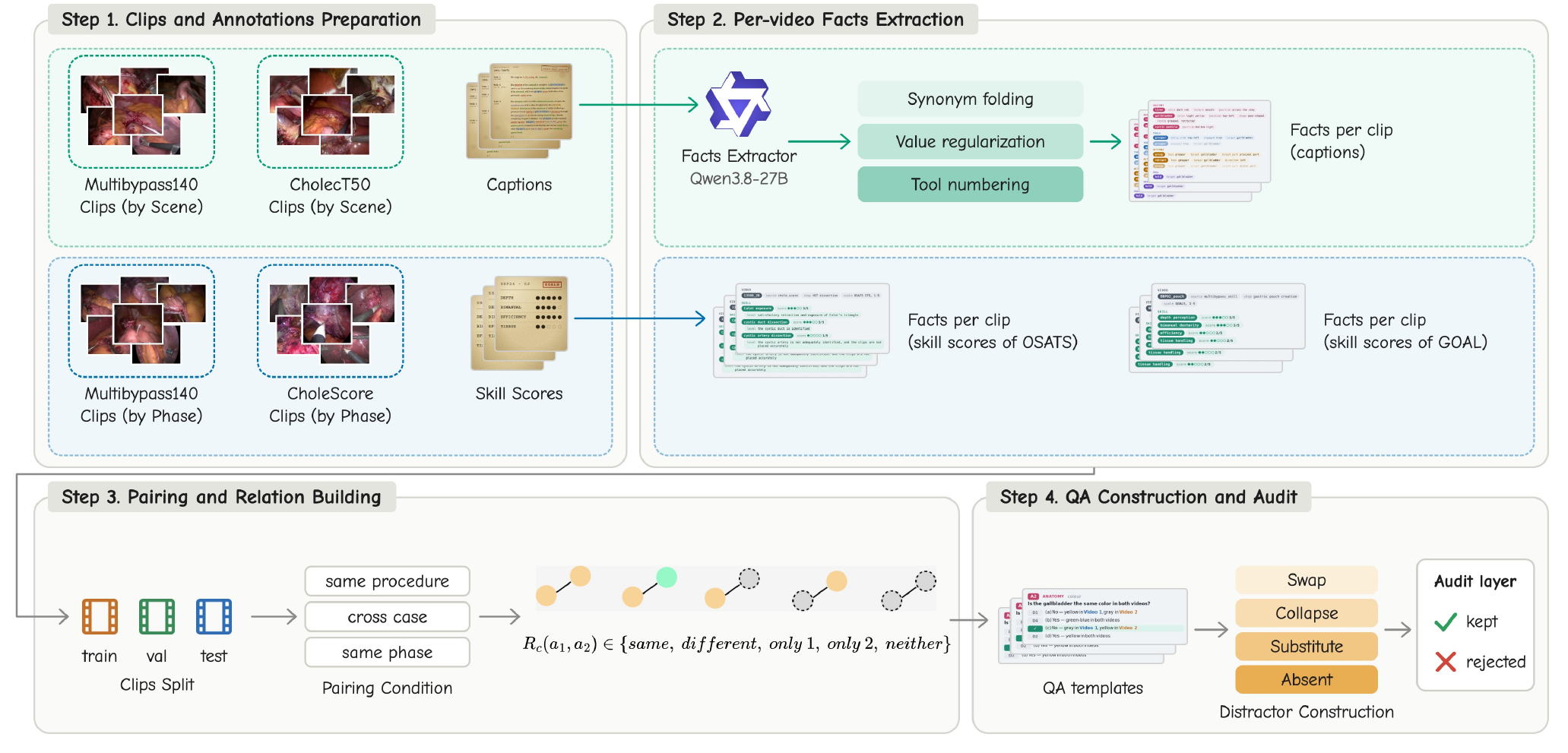}
\caption{An overview of EndoDiffVQA construction pipline.}\label{fig:pipeline}
\end{figure}

Both kinds of annotation are handled by one formulation. Each clip $v$ carries one source annotation
$\mathrm{ann}(v)$, its scene caption or its expert skill score. A fact extractor $f_c$ reads the annotation and
produces the \emph{per-clip fact} $a=f_c\big(\mathrm{ann}(v)\big)$ for an attribute $c$ of clip $v$; for example,
for the attribute $c = gallbladder\ state $, the fact $a$ may be \emph{porcelain}. A pair of clips is then described by a \emph{relation} $r=R_c(a_1,a_2)$, where $R_c$ is a deterministic comparison rule for attribute $c$, and a question template turns $(c,r,a_1,a_2)$ into a stem $q$, an option set $\mathcal{O}$ and the correct answer
$y^\star$, the letter of the option that states $r$. The same facts $a_1,a_2$ later supervise the per-video
reasoning of our model (Section~\ref{sec:ptc}). Fig.~\ref{fig:pipeline} shows the our pipeline of four stages, which we summarise here; Supplementary Section~S1 gives further details of the pipeline.

\textbf{Stage 1: Clip and annotation preparation.} \emph{Captions} describe single-scene clips of seconds to tens
of seconds and support Anatomy, Tool and Action questions. We use the captions of CholecT50~\citep{nwoye2022rendezvous} from~\citep{li2026surgtemp} and
additionally curate captions of the same form for MultiBypass140 clips. \emph{Skill scores} rate phase-level clips of tens of minutes and support Skill questions. We adopt the OSATS-style~\citep{martin1997osats} skill annotations of CholeScore~\citep{sharma2025early} and newly annotate MultiBypass140 with GOALS~\citep{vassiliou2005goals} skill
scores.

\textbf{Stage 2: Per-video facts extraction.} For captions, $f_c$ is a language model (Qwen3.8-27B) that fills a
fixed schema with one slot per attribute (visibility, colour, position and state of each structure; entry side,
engagement and target of each instrument; verb, target and part of each action; the surgeon's goal), followed by
synonym folding, value regularisation and removal of tool numbering so that facts are comparable across clips. For skill, the expert score is the fact, $a=s$.

\textbf{Stage 3: Pairing and relation building.} Clips are split by case into train, validation and test before
pairing. Two clips are paired only if they share the procedure and surgical phase and come from different cases. For every attribute stated in at least one clip, the pair receives
\begin{equation}
  R_c(a_1,a_2)\in\{\textsc{same},\ \textsc{different},\ \textsc{only}_1,\ \textsc{only}_2,\ \textsc{neither}\}
  \label{eq:relation}
\end{equation}
for captions, where \textsc{only}$_k$ and \textsc{neither} describe presence; for skill, with $\Delta=s_1-s_2$,
\textsc{same} if $\Delta=0$ and \textsc{different} with the sign of $\Delta$ otherwise.

\textbf{Stage 4: QA construction and audit.} Each relation is rendered by one of 13 templates (Supplementary
Table~S4) into a stem and four options. Every option opens with a verdict, \emph{Yes} or \emph{No}, followed by what
each clip shows, so a model must decide both whether the clips agree and which clip is which. The three
distractors are drawn from four named error types: \textbf{swap} (the correct facts on the wrong clips),
\textbf{collapse} (the opposite verdict), \textbf{substitute} (one side replaced by another value of the attribute)
and \textbf{absent} (a value neither clip shows). For quality control, items are sampled under per-template quotas
and relation mixes, and answer letters and question subjects are balanced so that neither predicts the answer. With clinicians, we reviewed the templates and pipeline and audited sampled items over several rounds until no new errors surfaced; the errors found and the resulting interventions are detailed in
Supplementary Table~S5.

The resulting dataset contains 10{,}939 items over 1{,}762 clips; example QA pairs and detailed statistics are
given in Supplementary Section~S2.

\runinheadings
\subsection{Modeling}\label{sec:modeling}

\subsubsection{Overall architecture}\label{sec:arch}

Each dataset item consists of two surgical videos $V_1$ and $V_2$, a comparative question $q$ and an option
set $\mathcal{O}=\{o_a,o_b,o_c,o_d\}$ with a single correct answer letter $y^\star\in\{a,b,c,d\}$. Our model builds on a
vision-language backbone Qwen3.5 made of three components: a vision transformer $E$, a merger $M$ that projects visual
tokens into the language embedding space, and an autoregressive language model $F_\theta$.

From each video we sample $T$ frames uniformly. The vision encoder turns them into patch tokens
$Z_k=E_k(V_k)\in\mathbb{R}^{N_k\times d_v}$, and the merger concatenates each $2\times 2$ group of neighbouring
patches and projects it to the language width $d$, giving $X_k=M_k(Z_k)\in\mathbb{R}^{(N_k/4)\times d}$, for
$k\in\{1,2\}$. The subscript $k$ marks the \emph{video slot}, the position of the video in the prompt.

\begin{figure}[t]
\centering
\includegraphics[width=\textwidth]{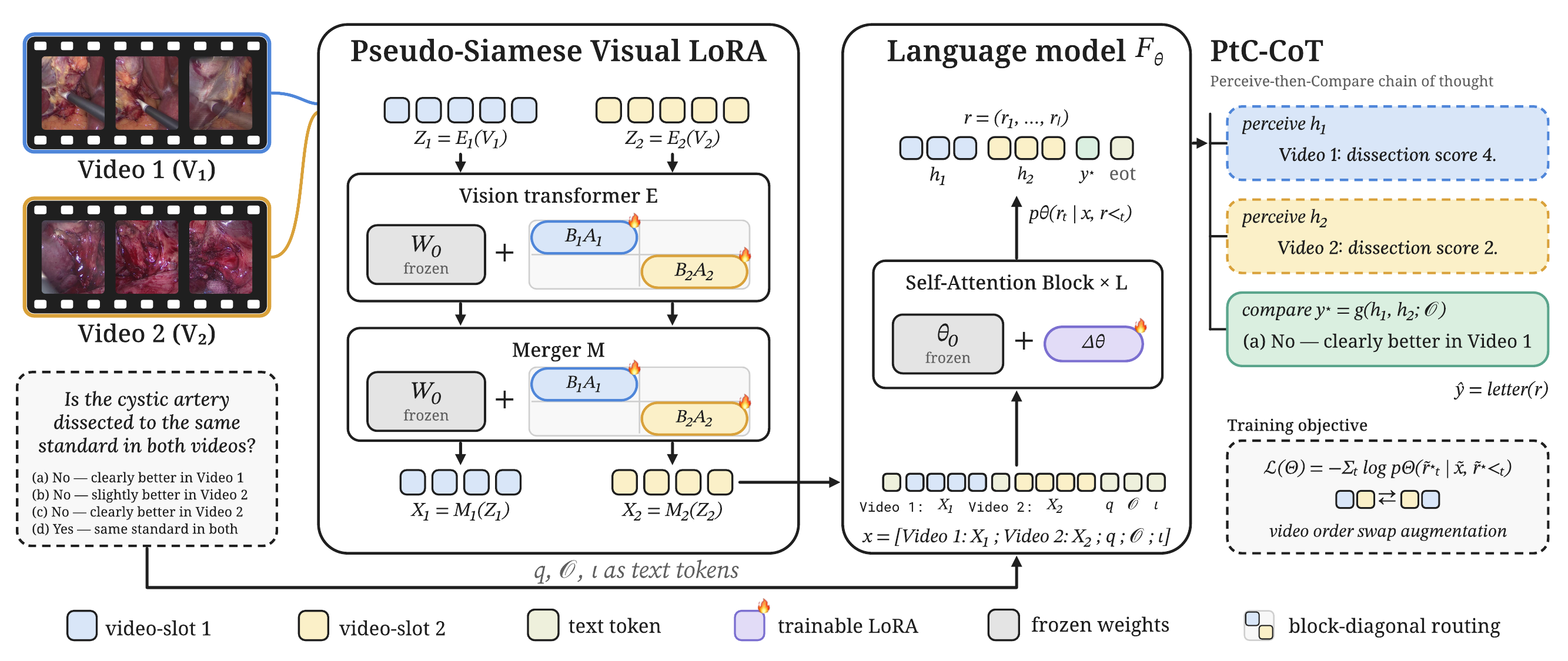}
\caption{An overview of EndoComparer model architecture.}\label{fig:model}
\end{figure}

The language model reads the interleaved sequence
\begin{equation}
  \mathbf{x}=\big[\,\texttt{Video 1:}\;X_1;\;\texttt{Video 2:}\;X_2;\;q;\;\mathcal{O};\;\iota\,\big],
  \label{eq:input}
\end{equation}
where $\iota$ is the instruction, and generates a response $\mathbf{r}=(r_1,\dots,r_L)$ token by token,
\begin{equation}
  p_\theta(\mathbf{r}\mid\mathbf{x})=\prod_{t=1}^{L}p_\theta\big(r_t\mid\mathbf{x},\mathbf{r}_{<t}\big).
  \label{eq:ar}
\end{equation}
The response is a Perceive-then-Compare chain of thought (Section~\ref{sec:ptc}) that first states what each video
shows and ends with the selected answer letter, $\hat{y}=\mathrm{letter}(\mathbf{r})$. Fig.~\ref{fig:model} gives an
overview of the model.

\subsubsection{Pseudo-Siamese Visual LoRA}\label{sec:psvl}

Comparative questions ask the model to bind each observation to the correct video. A Siamese visual encoder, which
applies identical weights to both inputs, leaves this binding entirely to the language model. We instead give each
video slot its own low-rank adaptation of the vision encoder and the merger: the two branches share the frozen
backbone weights and the architecture, but not the adaptation, which makes the visual pathway
\emph{pseudo-Siamese}.

Let $W_0\in\mathbb{R}^{d_\mathrm{out}\times d_\mathrm{in}}$ be a frozen linear layer of the visual pathway: the
query-key-value and output projections and the two MLP layers of every vision transformer block, and the two
linear layers of the merger. For slot $k$ we attach a pair of low-rank matrices
$A_k\in\mathbb{R}^{\rho\times d_\mathrm{in}}$ and $B_k\in\mathbb{R}^{d_\mathrm{out}\times\rho}$ with rank
$\rho\ll\min(d_\mathrm{in},d_\mathrm{out})$:
\begin{equation}
  W_k = W_0 + \frac{\alpha}{\rho}\,B_k A_k,\qquad k\in\{1,2\},
  \label{eq:slotlora}
\end{equation}
where $\alpha$ is the LoRA scaling factor. Collecting the adapters of all adapted layers in
$\Delta\phi_k=\{A_k,B_k\}$, the two visual branches are
\begin{equation}
  E_k = E(\,\cdot\,;\phi_0+\Delta\phi_k),\qquad M_k = M(\,\cdot\,;\phi_0+\Delta\phi_k),
  \label{eq:branches}
\end{equation}
with shared frozen weights $\phi_0$.

\paragraph{Block-diagonal routing} Both videos are encoded in one packed sequence
$Z=[Z_1;Z_2]\in\mathbb{R}^{(N_1+N_2)\times d_v}$. Self-attention in the vision transformer is restricted to a
block-diagonal mask over the two videos, so every block acts on each video independently. Each adapted layer
therefore routes its input row $z_i$ by the slot index
\begin{equation}
  s(i)=\begin{cases}1, & i\le N_1,\\ 2, & i>N_1,\end{cases}
  \qquad
  W_0 z_i+\frac{\alpha}{\rho}\,B_{s(i)}A_{s(i)}\,z_i,
  \label{eq:routing}
\end{equation}
with the boundary moved to $N_1/4$ in the merger. A single forward pass thus computes exactly the two branches of
Eq.~\eqref{eq:branches}, at the cost of one encoder pass.

\subsubsection{Perceive-then-Compare CoT}\label{sec:ptc}

A comparative answer is a function of two single-video facts: whether the clip applier is visible in each video,
how many instruments each video shows, or which skill score each surgeon reaches. Perceive-then-Compare CoT
(PtC-CoT) makes this decomposition explicit in the output: the model first perceives each video separately, then
compares.

For an item of attribute type $c$, the per-video facts $a_k=f_c(\mathrm{ann}(V_k))$ of Section~\ref{sec:dataset} are
known from dataset construction. We render each as a short perception statement,
\begin{equation}
  h_k=\pi_c(a_k),\qquad k\in\{1,2\},
  \label{eq:halves}
\end{equation}
where $\pi_c$ renders the fact as text. The correct letter is a deterministic comparison of the two statements, $y^\star=g_c(h_1,h_2;\mathcal{O})$, where $g_c$
applies the relation rule $R_c$ of Section~\ref{sec:dataset} to the two facts and returns the letter of the option
in $\mathcal{O}$ that states $R_c(a_1,a_2)$; this is exactly how $y^\star$ was assigned when the item was built. The target
response chains the two perception steps and the comparison:
\begin{equation}
  \mathbf{r}^\star=\texttt{Video 1: }h_1\texttt{. Video 2: }h_2\texttt{. (}y^\star\texttt{)}.
  \label{eq:target}
\end{equation}
Under Eq.~\eqref{eq:ar} this factorises the answer as
\begin{equation}
  p_\theta(\mathbf{r}^\star\mid\mathbf{x})=
  \underbrace{p_\theta(h_1\mid\mathbf{x})\,p_\theta(h_2\mid\mathbf{x},h_1)}_{\text{perceive}}\;
  \underbrace{p_\theta(y^\star\mid\mathbf{x},h_1,h_2)}_{\text{compare}},
  \label{eq:factor}
\end{equation}
so the letter is conditioned on the model's own per-video statements. Combined with the slot-specific encoders of
Section~\ref{sec:psvl}, each perception step $h_k$ is grounded in the visual tokens $X_k$ of its own branch. 
At inference we decode greedily with instruction $\iota$ asking the model to state what each video shows before the
letter, and read $\hat{y}$ from the final parenthesised letter.

\subsubsection{Training objective}\label{sec:objective}

The vision and merger weights $\phi_0$ and the language model weights $\theta_0$ stay frozen. We train the two slot
adapters $\Delta\phi_1,\Delta\phi_2$ together with a LoRA adapter $\Delta\theta$ on the attention, MLP and linear
recurrent projections of the language model, $\Theta=\{\Delta\theta,\Delta\phi_1,\Delta\phi_2\}$.

\paragraph{Order augmentation} With probability $\tfrac{1}{2}$ the videos of a training instance are exchanged,
$(V_1,V_2)\mapsto(V_2,V_1)$, which exchanges the perception statements $(h_1,h_2)\mapsto(h_2,h_1)$ and maps the
answer through the swap map $\sigma$ (e.g.\ \emph{only in Video 1} $\leftrightarrow$ \emph{only in Video 2},
\emph{better in Video 1} $\leftrightarrow$ \emph{better in Video 2}, while symmetric answers such as \emph{both}
or \emph{same} are fixed). Because slot $k$ is defined by position, every clip is seen by both visual branches over
training.

\paragraph{Loss} With $\tilde{\mathbf{x}}$ and $\tilde{\mathbf{r}}^\star$ the augmented input and PtC-CoT target, we
minimise the negative log-likelihood of the response tokens only, including the end-of-turn token:
\begin{equation}
  \mathcal{L}(\Theta)=-\,\mathbb{E}_{(\tilde{\mathbf{x}},\tilde{\mathbf{r}}^\star)}
  \sum_{t=1}^{|\tilde{\mathbf{r}}^\star|}\log p_{\Theta}\big(\tilde{r}^\star_t\mid\tilde{\mathbf{x}},
  \tilde{\mathbf{r}}^\star_{<t}\big).
  \label{eq:loss}
\end{equation}
The prompt tokens, including the visual tokens, are masked from the loss. Since $h_1$ and $h_2$ are supervised
directly, each slot adapter $\Delta\phi_k$ receives a dedicated perception signal for its own video in addition to
the gradient through the final comparison.

\blockheadings
\section{Experiments and results}\label{sec:results}
\runinheadings

\paragraph{Metrics}\label{sec:metrics}
Each option states one of the five relations of Eq.~\eqref{eq:relation}, grouped into \emph{Same}
$=\{\textsc{same},\textsc{neither}\}$ and \emph{Different} $=\{\textsc{different},\textsc{only}_1,\textsc{only}_2\}$.
We score each subset with three metrics and report their mean over the four subsets (4-subset macro).
\textbf{Detect} is the macro-F1 of the Same/Different decision; it charges both missed differences and false alarms.
\textbf{Specify} is the exact-option accuracy on the items whose correct answer is Different \emph{and} which the model also called Different; it probes how well a model specifies the difference (which video, which value, how much better). A subset in which a model calls fewer than 20 items Different is excluded from its macro and marked.
\textbf{Strict} is the macro-F1 over exact answers. With $\mathcal{F}_s$ the question families of subset $s$ and
$\mathcal{C}_f$ the exact answers that occur as gold in family $f$ (for presence
$\{\textsc{same},\textsc{only}_1,\textsc{only}_2,\textsc{neither}\}$; for skill
$\{\textsc{different}^{+},\textsc{different}^{-},\textsc{same}\}$, the sign giving the better video),
\begin{equation}
  \mathrm{Strict}_s=\frac{1}{|\mathcal{F}_s|}\sum_{f\in\mathcal{F}_s}\frac{1}{|\mathcal{C}_f|}
  \sum_{c\in\mathcal{C}_f}\frac{2\,\mathrm{TP}_{f,c}}{2\,\mathrm{TP}_{f,c}+\mathrm{FP}_{f,c}+\mathrm{FN}_{f,c}},
  \label{eq:strict}
\end{equation}
where $\mathrm{TP}_{f,c}$, $\mathrm{FP}_{f,c}$ and $\mathrm{FN}_{f,c}$ count the items of family $f$ in which
answer $c$ is chosen and correct, chosen but wrong, and correct but not chosen. Every answer weighs the same, so
collapsing to one answer scores low whichever answer it is.

\paragraph{Evaluation protocol}\label{sec:protocol}
All models receive the same input: 16 uniformly sampled frames per caption clip and 64 per skill clip, each at
most $784\times448$ pixels, from one shared frame cache, followed by the question, the four lettered options and
the instruction. Decoding is greedy. Fine-tuned models are trained on train\,+\,validation for one epoch, and the final checkpoint is evaluated on
the test split. Results are the mean of three training runs with different seeds.

\paragraph{Implementation details}\label{sec:impl}
Our model builds on Qwen3.5-9B. The language-model LoRA (rank 32, $\alpha=64$, dropout 0.05) covers the attention,
MLP and linear-recurrent projections of all 32 layers; the two slot adapters cover the four linear layers of each
of the 27 vision transformer blocks and the two merger layers, 110 modules per slot, with slot~2 initialised from
slot~1's $A$ and $B=0$ so that both branches start from the backbone. Training uses AdamW with learning rate
$10^{-4}$, cosine decay, 3\,\% warm-up and no weight decay, for one epoch, at an effective batch of 16 on two
H100 GPUs, about 3.5\,h.

\paragraph{Comparison with baselines}\label{sec:main-results}

\begin{table}[t]
\caption{Results on the EndoDiffVQA test split (2{,}300 items, \%). Detect, Specify and Strict are 4-subset
macros (see Metrics); the last four columns give Strict per subset. $^\dagger$: Specify macro over
fewer than four subsets, because the model called fewer than 20 items Different in the others; not highlighted.
Best per group in bold; the fine-tuned group includes ours.}\label{tab:baselines}
\setlength{\tabcolsep}{3pt}
\begin{tabular}{@{}lccc|cccc@{}}
\toprule
 & \multicolumn{3}{c|}{4-subset macro} & \multicolumn{4}{c}{Strict per subset} \\
Model & Detect & Specify & Strict & \begin{tabular}[c]{@{}c@{}}Chole-\\Cap\end{tabular} & \begin{tabular}[c]{@{}c@{}}LRYGB-\\Cap\end{tabular} & \begin{tabular}[c]{@{}c@{}}Chole-\\Skill\end{tabular} & \begin{tabular}[c]{@{}c@{}}LRYGB-\\Skill\end{tabular} \\
\midrule
\multicolumn{8}{@{}l}{\emph{Zero-shot}} \\
Qwen3.5-9B~\citep{qwen35}        & 49.8\sd{.5} & 57.3\sd{.4} & 35.4\sd{.9} & 41.3\sd{.3} & 34.4\sd{.7} & 31.6\sd{.6} & 34.3\sd{.3} \\
Qwen3.8-27B~\citep{qwen38}       & 52.8\sd{.7} & 58.9$^\dagger$\sd{.2} & 35.2\sd{.6} & 38.7\sd{.3} & 33.8\sd{.3} & \textbf{44.7}\sd{.6} & 23.6\sd{1.0} \\
InternVL3.5-8B~\citep{wang2025internvl35}    & 49.7\sd{.3} & 54.5$^\dagger$\sd{.4} & 30.6\sd{.8} & 44.7\sd{1.1} & 25.5\sd{.8} & 28.6\sd{.6} & 23.6\sd{1.2} \\
InternVL3.5-30B~\citep{wang2025internvl35}   & 56.6\sd{.2} & 57.1\sd{1.1} & 37.2\sd{.5} & 44.5\sd{.3} & 32.0\sd{.3} & 36.3\sd{.5} & 36.1\sd{1.0} \\
SurgVidLM~\citep{wang2025surgvidlm}         & 51.5\sd{.4} & 49.2$^\dagger$\sd{.8} & 32.1\sd{.8} & 32.5\sd{.6} & 24.9\sd{.7} & 39.2\sd{.3} & 32.0\sd{.3} \\
GPT-5.5~\citep{openai2026gpt55}           & 49.5\sd{.4} & 56.3\sd{.9} & 32.9\sd{.6} & 45.8\sd{.5} & 25.9\sd{.8} & 35.2\sd{.7} & 24.7\sd{.5} \\
Gemini 3.6 Flash~\citep{google2026gemini36}  & \textbf{59.1}\sd{1.0} & \textbf{75.1}\sd{.9} & \textbf{46.9}\sd{.4} & \textbf{53.9}\sd{.8} & \textbf{46.9}\sd{.7} & 43.1\sd{1.1} & \textbf{43.7}\sd{.9} \\
\midrule
\multicolumn{8}{@{}l}{\emph{Fine-tuned}} \\
InternVL3.5-8B~\citep{wang2025internvl35}    & 61.6\sd{.5} & 85.3$^\dagger$\sd{1.2} & 47.0\sd{.3} & 66.3\sd{.6} & 51.7\sd{1.0} & 46.4\sd{.4} & 23.6\sd{.7} \\
SurgVidLM~\citep{wang2025surgvidlm}         & 60.0\sd{.2} & 88.3$^\dagger$\sd{.9} & 45.1\sd{1.0} & 62.6\sd{.8} & 52.2\sd{1.1} & 38.0\sd{.5} & 27.5\sd{.9} \\
SurgTEMP~\citep{li2026surgtemp}  & 60.9\sd{.8} & 88.9$^\dagger$\sd{.8} & 46.8\sd{.7} & 64.3\sd{1.0} & 52.9\sd{1.1} & 42.6\sd{.7} & 27.5\sd{.9} \\
SurgViVQA~\citep{drago2025surgvivqa}         & 52.4\sd{.3} & 37.5\sd{.9} & 32.6\sd{.8} & 32.9\sd{1.2} & 33.1\sd{1.0} & 36.9\sd{.5} & 27.4\sd{.6} \\
PeskaVLP~\citep{yuan2024peskavlp}   & 61.3\sd{.9} & 63.3$^\dagger$\sd{.2} & 44.6\sd{.7} & 53.9\sd{.4} & 44.8\sd{.3} & 47.7\sd{.3} & 31.9\sd{1.0} \\
SurgCLIP~\citep{perez2025surglavi}   & 61.4\sd{.3} & 57.3\sd{.4} & 43.5\sd{.6} & 54.0\sd{1.1} & 45.4\sd{.3} & 47.5\sd{.6} & 27.2\sd{.7} \\
Qwen3.5-9B~\citep{qwen35}        & 63.6\sd{1.1} & 77.6\sd{1.0} & 48.3\sd{1.1} & 64.6\sd{.5} & 49.4\sd{.6} & 43.3\sd{.6} & 31.1\sd{1.1} \\
\midrule
\quad w/o PSVL & 64.0\sd{1.2} & 78.0\sd{.4} & 49.5\sd{.4} & 65.3\sd{.4} & 51.6\sd{.4} & 44.9\sd{.7} & 33.4\sd{.8} \\
\quad w/o PtC-CoT & 65.6\sd{.5} & 80.3\sd{.2} & 50.9\sd{.6} & 68.3\sd{.6} & 52.5\sd{.8} & 43.3\sd{1.2} & 32.6\sd{.9} \\
EndoComparer (ours) & \textbf{67.7}\sd{.8} & \textbf{83.5}\sd{.6} & \textbf{54.0}\sd{.4} & \textbf{69.6}\sd{.5} & \textbf{54.5}\sd{1.2} & \textbf{51.9}\sd{.8} & \textbf{40.1}\sd{1.1} \\
\bottomrule
\end{tabular}
\end{table}

Table~\ref{tab:baselines} shows that EndoDiffVQA is hard for current models. Zero-shot, open models reach only
28.0--37.2 Strict, and the surgical VLM (SurgVidLM) score below the general ones. Gemini 3.6 Flash is the
strongest zero-shot model (46.9 Strict), close to the fine-tuned Qwen3.5-9B (48.3). Fine-tuning raises Strict by
13--19 points for the models evaluated both ways, but the gain comes from the caption subsets: on LRYGB-Skill the
fine-tuned models stay at 23.6--31.9, no better than zero-shot. The high Specify of fine-tuned InternVL3.5, SurgVidLM and SurgTEMP is computed over fewer subsets, because these models seldom answer Different. EndoComparer is best on every column of its group except Specify. It improves over its fine-tuned Qwen3.5-9B backbone by 4.1 Detect, 5.9 Specify and 5.7 Strict, and over Gemini 3.6 Flash by 7.1 Strict, with the largest gains on the skill subsets (+8.6 and +9.0 Strict on Chole-Skill and LRYGB-Skill).

\paragraph{Ablation study}\label{sec:ablation}
The last three rows of Table~\ref{tab:baselines} start from the fine-tuned Qwen3.5-9B, which uses a Siamese vision
LoRA and the letter-only target. Each component helps on its own: PtC-CoT alone (w/o PSVL) adds 1.2 Strict and PSVL
alone (w/o PtC-CoT) adds 2.6, the latter almost entirely on the caption subsets. Combining them adds 5.7 Strict, and the gain on the skill subsets is larger than either component gives alone: Chole-Skill rises from 44.9 and 43.3 to 51.9, and LRYGB-Skill from 33.4 and 32.6 to 40.1. The two components are complementary: the per-video statements of PtC-CoT are most useful when each statement comes from its own visual branch.

\paragraph{Robustness analysis}\label{sec:swap}
We conduct the swap control experiment to measure the video confusion problem and the models' robustness directly.
Table~\ref{tab:swap} re-evaluates the test items with Video~1 and Video~2 exchanged, scored against the swapped
answer. The drop in Strict quantifies the failures after the swap. The fine-tuned VLMs lose 5--8 points, confirming
the failure: they perceive the difference and attach it to the wrong video. EndoComparer loses 0.6, which indicates
that distinguishing the two videos in the visual pathway (PSVL) and in the reasoning (PtC-CoT) mitigates the confusion that a shared pathway and a letter target leave in place.

\begin{table}[!ht]
\caption{Robustness to video order: Strict on the EndoDiffVQA test split (\%, 4-subset macro). \emph{Full} is the release
order; \emph{Swap} exchanges Video~1 and Video~2 and scores against the swapped answer. $\Delta$ is Swap $-$ Full.}
\label{tab:swap}
\footnotesize\setlength{\tabcolsep}{2.5pt}
\begin{tabular}{@{}l|ccc|cccccc@{}}
\toprule
 & \multicolumn{3}{c|}{Zero-shot} & \multicolumn{6}{c}{Fine-tuned} \\
 & \begin{tabular}[c]{@{}c@{}}Qwen3.8\\27B\end{tabular}
 & \begin{tabular}[c]{@{}c@{}}InternVL\\3.5-30B\end{tabular}
 & \begin{tabular}[c]{@{}c@{}}Surg-\\VidLM\end{tabular}
 & \begin{tabular}[c]{@{}c@{}}InternVL\\3.5-8B\end{tabular}
 & \begin{tabular}[c]{@{}c@{}}Surg-\\VidLM\end{tabular}
 & \begin{tabular}[c]{@{}c@{}}Surg-\\TEMP\end{tabular}
 & \begin{tabular}[c]{@{}c@{}}Surg-\\ViVQA\end{tabular}
 & \begin{tabular}[c]{@{}c@{}}Qwen3.5\\9B\end{tabular}
 & \begin{tabular}[c]{@{}c@{}}Endo-\\Comparer\end{tabular} \\
\midrule
Full & 35.2 & 37.2 & 32.1 & 47.0 & 45.1 & 46.8 & 32.6 & 48.3 & \textbf{54.0} \\
Swap & 33.1 & 34.5 & 27.8 & 41.6 & 38.3 & 39.2 & 27.4 & 42.5 & \textbf{53.6} \\
$\Delta$ & $-$2.1 & $-$2.7 & $-$4.3 & $-$5.4 & $-$6.8 & $-$7.6 & $-$5.2 & $-$5.8 & $\mathbf{-0.6}$ \\
\bottomrule
\end{tabular}
\end{table}

\section{Discussion}\label{sec:discussion}

The results support the view that comparison is not single-video understanding applied twice. Fine-tuning a
Siamese backbone teaches it the caption subsets but leaves skill comparison near the zero-shot level, and the swap
control shows that many of its correct answers depend on which video sits in which slot. Both components of
EndoComparer address attribution rather than perception, and their combination helps most on skill, where the two
clips show the same step of the same procedure and differ only in how it is performed. Several limitations
apply: the benchmark compares two videos only and inherits the coverage and inter-rater noise of its source
captions and scores; multiple choice measures whether a model selects the correct relation, not whether it can
describe a difference freely; Clinically, the per-video
statements are a first step towards feedback that names what differs and in which video, although at the present
level the model suits ranking and triage of recorded cases rather than unsupervised feedback. More qualitative results are given in Supplementary material. 

\section{Conclusion}\label{sec:conclusion}

We introduced EndoDiffVQA, a benchmark for comparative surgical video question answering whose coverage ranges from
fundamental to complex aspects of surgery: anatomy, tools, actions and skill. A hierarchical metric system
evaluates a model's ability to detect and to specify a difference while preventing the exploitation of option
patterns. We further proposed EndoComparer, with dedicated designs in the visual pathway (PSVL) and in the reasoning
(PtC-CoT) that mitigate the video confusion dilemma; the two components are complementary and together set a strong
baseline over current generalist and surgical VLMs.

\backmatter

\section*{Declarations}
\bmhead{Funding} 
This work has received funding from the European Union (ERC, CompSURG, 101088553).
Views and opinions expressed are however those of the authors only and do not necessarily reflect those of the European Union or the European Research Council. Neither the European Union nor the granting authority can be held responsible for them.
This work was also partially supported by French state funds managed by the ANR under Grant ANR-10-IAHU-02 (IHU Strasbourg) and by the Interdisciplinary Thematic Institute HealthTech (ITI 2021-2028 program of the University of Strasbourg, CNRS and Inserm) via the IdEx Unistra (ANR-10-IDEX-0002) and SFRI (STRATUS project, ANR-20-SFRI-0012).
This work was also granted access to the servers/HPC resources managed by CAMMA, IHU Strasbourg and Unistra Mesocentre.
\bmhead{Conflict of interest} Pietro Mascagni and Nicolas Padoy are co-founders and shareholders of Scialytics.
\bmhead{Ethical approval} CholeScore is part of a study approved by University of Strasbourg’s Ethics Committee (ID F20200730144229); the data was obtained with patient consent. The other datasets are public.
\bmhead{Data, code and/or material availability}
Code is public; the annotations we curated on the public datasets will be released upon acceptance.

\bibliography{references}

\clearpage
\setcounter{section}{0}
\setcounter{table}{0}
\setcounter{figure}{0}
\setcounter{equation}{0}
\renewcommand{\thesection}{S\arabic{section}}
\renewcommand{\thetable}{S\arabic{table}}
\renewcommand{\thefigure}{S\arabic{figure}}
\renewcommand{\theequation}{S\arabic{equation}}
\renewcommand{\theHsection}{S\arabic{section}}
\renewcommand{\theHtable}{S\arabic{table}}
\renewcommand{\theHfigure}{S\arabic{figure}}
\renewcommand{\theHequation}{S\arabic{equation}}
\setcounter{secnumdepth}{3}
\section*{Supplementary material}
\addtocontents{toc}{\protect\setcounter{tocdepth}{3}}
\tableofcontents
\section{Dataset construction details}\label{supp:sec:dataset}

This section expands the four construction stages summarised in Section~2.1 of the main paper, one subsection per
stage. Dataset statistics follow in Section~\ref{sec:stats}.

\subsection{Stage 1: Clip and annotation preparation}\label{sec:stage1}

\emph{Captions} describe short clips, each covering a single scene and
lasting from seconds to tens of seconds, and support questions about Anatomy, Tool and Action. \emph{Skill
proficiency scores} rate how well a surgical step is performed and support Skill questions: \choleskill is built
from the OSATS-style criteria of CholeScore, scored 1--5 per phase video, and \lrygbskill from the GOALS domains
(depth perception, bimanual dexterity, efficiency, tissue handling) rated on step segments of MultiBypass140. These
phase-level clips last tens of minutes on average; Table~\ref{tab:duration} in Section~\ref{sec:stats} gives the
clip duration statistics of the two annotation types.

\subsection{Stage 2: Per-video fact extraction}\label{sec:slots}

This subsection details the fact extractor $f_c$ of Stage~2 for the caption lane. Each clip caption is
turned into one JSON record with four parts: \texttt{anatomy}, \texttt{tools}, \texttt{actions} and \texttt{goal}.
The extractor reads the caption only, never the video, so a slot is filled only when the caption states it.

\subsubsection{Schema}

Table~\ref{tab:slots} lists every slot, its value type and the question templates that read it.
Slots with a closed vocabulary are enforced at decoding time; free-text slots copy the caption's words and are
normalised afterwards. A1--A4, T1--T4 and C1--C3 are the Anatomy, Tool and Action templates of
Table~\ref{tab:templates}.

\begin{table}[!ht]
\caption{Slot schema of the per-clip record: one \texttt{anatomy} entry per structure, one \texttt{tools} entry per instrument instance, one \texttt{actions} entry per instrument--tissue interaction, and one \texttt{goal}. Region: one of \emph{left, right, top, bottom, center, top-left, top-right,
bottom-left, bottom-right, background, across the view}. Slots that are extracted but feed no released template
(texture, shape, exposure, movement direction) are omitted.}\label{tab:slots}
\footnotesize
\begin{tabular}{@{}lL{2.0cm}L{6.6cm}L{2.0cm}@{}}
\toprule
Record & Slot & Value & Templates \\
\midrule
\texttt{anatomy}  & \texttt{entity}         & canonical lower-case noun (``gallbladder'', ``cystic pedicle'') & A1--A3 \\
                  & \texttt{visible}        & boolean; false only if the caption says the structure is not seen & A1 \\
                  & \texttt{color}          & caption's colour words, or null & A2 \\
                  & \texttt{position}       & region, or null & A3 \\
                  & \texttt{states}         & subset of \emph{bleeding, dissected, cut, clipped, porcelain, retracted, grasped,
                                              cauterized, coagulated, stapled, sutured, inflamed, fatty, distended, exposed} & A4 \\
\midrule
\texttt{tools}    & \texttt{tool}           & canonical instrument name, numbering stripped & T1, T3, T4 \\
                  & \texttt{visible}        & boolean; false inside ``(Not seen on this clip \dots)'' & T1 \\
                  & \texttt{entry\_side}    & region, or null & T2 \\
                  & \texttt{engaged}        & true / false / null (acting on tissue, idle, unclear) & T3, T4 \\
                  & \texttt{target}         & structure acted on, or null & T3, T4 \\
\midrule
\texttt{actions}  & \texttt{tool}           & instrument, numbering stripped, or null & C1 \\
                  & \texttt{verb}           & one of 25 verbs: \emph{grasp, hold, retract, lift, push, pull, reach, dissect, cut, clip,
                                              coagulate, cauterize, divide, irrigate, suction, clean, staple, suture, test, insert,
                                              withdraw, pack, extract, expose, other} & C1 \\
                  & \texttt{target}         & structure acted on, or null & C1, C2 \\
                  & \texttt{target\_part}   & part named (``proximal side'', ``fundus''), or null & C2 \\
                  & \texttt{visible}        & boolean; false if described as not seen on this clip & C1, C2 \\
\midrule
\texttt{goal}     & \texttt{verb}, \texttt{target} & surgeon's stated purpose; verb from the action list, or null & C3 \\
\bottomrule
\end{tabular}
\end{table}

\subsubsection{Prompt and decoding}

The extractor is Qwen3.8-27B, run text-only with vLLM. The system message gives the
schema and one rule per slot: copy the caption's own words, never infer, and use null when the caption is silent.
It also fixes four conventions that the audit showed to matter: instrument numbering is stripped (``grasper~1'' and
``grasper~2'' become two \emph{grasper} entries); specimen bag, trocar, compress, suture thread and needle are tools,
never anatomy; facts inside ``(Not seen on this clip \dots)'' are kept with \texttt{visible}~=~false; and a state is
recorded only when the caption asserts it visually (``dissected'' does not imply ``bleeding''). Two worked examples
follow as user/assistant turns, one numbered CholecT50 caption and one tiered MultiBypass140 caption, and the last
user turn holds the clip's surgical phase and caption. Decoding is greedy and is constrained
by the JSON schema, so every closed slot can only take a value from its vocabulary. Outputs that fail to parse are
retried once with sampling (temperature 0.7, top-$p$ 0.8, top-$k$ 20).

\subsubsection{Example}

Table~\ref{tab:slotex} shows an abbreviated CholecT50 caption and the slots extracted from it.

\begin{table}[!ht]
\caption{Slot extraction on an abbreviated CholecT50 caption (one entry per record shown). The numbered graspers
become two \emph{grasper} entries, and the grasp narrated inside ``(Not seen \dots)'' is kept as an unseen action.}
\label{tab:slotex}
\footnotesize
\begin{tabular}{@{}p{11.6cm}@{}}
\toprule
\emph{We can see a blue-green structure with thin red blood vessels on the top-left side of the camera, which is a
porcelain gallbladder. The proximal side of the gallbladder and its peritoneum are dissected [\dots] The grasper~2 is
seen on the left side of the camera. [\dots] (Not seen on this clip because of the camera's view: The grasper~1 is
grasping the distal side of the gallbladder.) The grasper~2 is retracting the proximal side of the gallbladder to the
left side of the camera. [\dots] The surgeon wants to dissect the gallbladder/cystic pedicle.} \\
\midrule
\begin{minipage}{11.6cm}
\begin{verbatim}
anatomy: {entity: gallbladder, visible: true, color: blue-green, position: top-left,
          states: [porcelain, dissected], texture/shape/exposure: null}
tools:   {tool: grasper, visible: true, entry_side: left, engaged: true,
          target: gallbladder}
actions: {tool: grasper, verb: retract, target: gallbladder, target_part: proximal side,
          direction: left, visible: true}
         {tool: grasper, verb: grasp, target: gallbladder, target_part: distal side,
          visible: false}
goal:    {verb: dissect, target: cystic pedicle}
\end{verbatim}
\end{minipage} \\
\bottomrule
\end{tabular}
\end{table}

\subsubsection{Post-processing}

The raw record then passes through deterministic steps (Table~\ref{tab:slotpost}).
Folding and regularisation make values comparable across clips; the caption-evidence guards re-read the caption and
remove slots the model filled without visual support. Most guards are the hard checks of Table~\ref{tab:audit}.

\begin{table}[!ht]
\caption{Deterministic post-processing of the extracted slots.}\label{tab:slotpost}
\footnotesize
\begin{tabular}{@{}L{3.0cm}L{8.6cm}@{}}
\toprule
Step & Rule \\
\midrule
\multicolumn{2}{@{}l}{\emph{Normalisation}} \\
Synonym folding & Surface variants map to one value: \emph{pinkish}$\to$\emph{pink}, \emph{upper left}$\to$\emph{top-left},
  \emph{clip applier}$\to$\emph{clipper}, \emph{gall bladder}$\to$\emph{gallbladder}, \emph{bleeds}$\to$\emph{bleeding};
  articles and punctuation are stripped. \\
Tool numbering & ``grasper~2''$\to$\emph{grasper}; any digit left in a tool name fails validation. \\
Object nouns & Specimen bag, trocar, compress, suture thread and needle filed under anatomy are moved to tools, and are
  never engaged and never an action's actor. \\
Entity families & Names one structure carries across the two corpora (\emph{omentum, mesentery, fat}) are kept but grouped,
  so absence (A1) must hold for the whole family. \\
Duplicates & Repeated entries are merged; slots of a structure marked not visible are cleared. \\
\midrule
\multicolumn{2}{@{}l}{\emph{Caption-evidence guards}} \\
Negation & A colour, texture or shape whose every mention sits in a negated clause is dropped
  (``no unified tubular structure''). \\
Release & A tool the caption says is releasing, resting on or loosening its target is not engaged with it. \\
Engagement & A tool is engaged with a target only through a visible action on it other than \emph{reach}. \\
Part vs.\ destination & ``retracting the gallbladder to its lateral part'' names where the tissue goes, not the part held;
  \texttt{target\_part} is cleared. \\
Entry side & A retractor side read off the liver's quadrant, with no side given for the retractor, is cleared. \\
Multiple targets & ``the grasper is reaching the X and the Y'' yields one action per target. \\
States from actions & A visible action adds the state it leaves (\emph{retract}$\to$\emph{retracted}); a transient state
  (\emph{retracted, grasped}) backed only by an unseen action is removed. \\
\midrule
\multicolumn{2}{@{}l}{\emph{Value regularisation}} \\
Goal map & The goal is decided once per distinct rationale sentence (448 in the corpus, 218 with a usable goal) rather
  than per clip, so two clips with the same sentence cannot disagree. A candidate goal is kept only if its target is the
  object of its verb in the sentence; targets naming an instrument, the camera, a spot or a fluid are set to null. \\
\bottomrule
\end{tabular}
\end{table}

\subsection{Stage 3: Pairing and relation building}\label{sec:stage3}

Clips are split into train, validation and test by case before any pair is formed.
Within a split, two clips are paired only under three conditions: they show the \emph{same procedure}, they come
from \emph{different cases}, and they cover the \emph{same surgical phase}, so that their facts are comparable. A cap
on how many pairs a clip may join (3--8 depending on the source) prevents a few clips from dominating.

\subsection{Stage 4: QA construction and audit}\label{sec:stage4}

\subsubsection{Templates and options}\label{sec:templates}

Every item is produced by one of 13 question templates, 11 for the caption lane and two for the skill lane
(Table~\ref{tab:templates}). A template reads one slot of the per-clip record of Section~\ref{sec:slots} from
each video (the fact), turns the two facts into a relation with a fixed rule, and fills a stem whose placeholders,
written \ph{entity}, \ph{tool}, \dots, name only the \emph{subject} of the comparison, never a value, so the
values can be read only from the videos. Figs.~\ref{fig:ex-anatomy}--\ref{fig:ex-skill} show one released item
per template.

\begin{table}[p]
\caption{The 13 question templates. Fact: what is read from each clip's record; Rule: how the two facts give the
relation of Eq.~(1) of the main paper, a pair that fits no case being skipped. Throughout, \textsc{only}$_k$
requires the subject to be safely absent from the other clip (no same-organ alias such as omentum / mesentery /
fat, no mention in its caption); overlapping values (``left'' vs.\ ``top-left'', ``yellow'' vs.\ ``yellow-pink'')
are never keyed as \textsc{different}; specimen bag, trocar and other objects never count as instruments. C1
action kinds: manipulate (grasp, hold, retract, lift, push, pull, expose), divide (cut, divide), energy
(coagulate, cauterize), clean (irrigate, suction, clean), and dissect, clip, staple, suture each on their
own.}\label{tab:templates}
\footnotesize
\setlength{\tabcolsep}{4pt}
\renewcommand{\arraystretch}{0.97}
\begin{tabular}{@{}p{0.45cm}L{2.8cm}L{4.0cm}L{4.9cm}@{}}
\toprule
ID & Fact per clip & Stem & Rule \\
\midrule
\multicolumn{4}{@{}l}{\textbf{Anatomy}}\\
A1 & structure visible? &
  Is the \ph{entity} visible in both videos? &
  \textsc{same} / \textsc{only}$_k$ / \textsc{neither} from the two flags; \textsc{neither}: a structure of the
  phase that neither caption mentions. \\
A2 & colour of the structure &
  Is the \ph{entity} the same colour in both videos? &
  \textsc{same} if equal, \textsc{different} if no colour word is shared, \textsc{only}$_k$ if absent from the
  other clip. \\
A3 & region of the view the structure is in &
  Is the \ph{entity} in the same part of the view in both videos? &
  As A2 over regions. \\
A4 & state of the structure (bleeding, dissected, clipped, \dots) &
  Is the \ph{entity} \ph{state} in both videos? &
  As A1 per (structure, state); \textsc{neither}: a state the structure shows elsewhere in the source. \\
\midrule
\multicolumn{4}{@{}l}{\textbf{Tool}}\\
T1 & instrument visible? &
  Is \ph{a tool} used in both videos? &
  As A1 over instruments. \\
T2 & side the instrument enters from &
  Does the \ph{tool} enter from the same side in both videos? &
  As A2 over sides. \\
T3 & number of instruments engaging the structure &
  Do both videos show the same number of instruments engaging the \ph{entity}? &
  \textsc{same} / more in Video~$k$ from the sign of $n_1-n_2$; \textsc{neither} when both counts are zero. \\
T4 & instrument acting on the structure? (reaching or releasing does not count) &
  Is the \ph{tool} actively engaged with the \ph{entity} in both videos? &
  As A1 from the two flags; \textsc{neither} only when the instrument is explicitly idle in both clips. \\
\midrule
\multicolumn{4}{@{}l}{\textbf{Action}}\\
C1 & seen (instrument, verb, tissue) triples, e.g.\ ``a hook dissects the gallbladder'' &
  Is the same interaction with the tissue shown in both videos? &
  \textsc{same} if a triple is shared; \textsc{different} only if no action of the same kind on the same organ
  occurs in both. \\
C2 & part of the structure held (grasp, hold, retract, lift, pull) &
  Does \ph{a tool} hold the same part of the \ph{entity} in both videos? &
  \textsc{same} / \textsc{different}. \\
C3 & surgeon's goal: verb and target, e.g.\ ``dissecting the gallbladder'' &
  Is the surgeon pursuing the same goal in both videos? &
  \textsc{same} if the verbs match and the targets share a structure, else \textsc{different}. \\
\midrule
\multicolumn{4}{@{}l}{\textbf{Skill}}\\
S1 & CholeScore score $s\in\{1,\dots,5\}$ of the phase video on one of five criteria (Calot exposure, cystic
  duct / artery dissection, clipping and transection, fossa dissection) &
  One stem per criterion, e.g.\ Is Calot's triangle exposed to the same standard in both videos? &
  $\Delta=s_1-s_2$: \textsc{same} if $\Delta=0$, \textsc{different} with the sign of $\Delta$ if
  $|\Delta|\ge 1$. Pairs share the criterion. \\
S2 & GOALS score $s\in\{1,\dots,5\}$ of a step segment ($\ge 2$\,min) on one of four domains (depth
  perception, bimanual dexterity, efficiency, tissue handling) &
  One stem per domain, e.g.\ Is two-handed coordination of the same standard in both videos during the
  \ph{step}? &
  As S1; pairs share the phase and the domain. \\
\bottomrule
\end{tabular}
\end{table}

\paragraph{Options and distractors.} Every option opens with a verdict, \emph{Yes} or \emph{No}, followed by what
each video shows, and the three distractors are of four named types: \textbf{D1 swap}, the correct facts attached
to the wrong videos, always present when the answer is asymmetric; \textbf{D2 collapse}, the opposite verdict;
\textbf{D3 substitute}, one side correct and the other replaced by a fake value; \textbf{D4 absent}, a value that
neither video shows. When more than three distractors are possible, the item keeps D1 and two of the others at
random. For presence, state, count and engagement (A1, A4, T1, T3, T4) the four options are always the four
relations in the template's wording, so the option set is identical whatever the answer. For the value templates
(A2, A3, T2, C1--C3) the options spell the values out; a fake value is drawn, frequency weighted, from the values
the same slot takes in other clips of the same source and surgical phase, and is never a value either clip attests
for the subject nor one overlapping with it (for colour and state, only values the \emph{same structure} takes
elsewhere, so a fake ``dark red gallbladder'' cannot be ruled out from anatomy alone). For skill the options are
``to the same standard in both videos'', ``clearly better in Video~1'', ``clearly better in Video~2'' and
``slightly better in Video~$k$''; the last is never correct. S1 also has a \emph{rubric} form, built only for
scores at the rubric's anchor levels 1, 3 and 5, whose options quote the CholeScore wording of two levels (``both
at `satisfactory retraction and exposure of Calot's triangle'\,'') with two Yes and two No options.

\subsubsection{Sampling quotas and answer balance}

To prevent a single template from dominating the dataset,
items are sampled under per-template quotas of 500/100/150 train/validation/test items per caption template, with a
minimum share of MultiBypass140 items, and target relation mixes of 50\,\% \textsc{different}, 25\,\% \textsc{same},
20\,\% \textsc{only} and 5\,\% \textsc{neither} for captions and 50/50 \textsc{same}/\textsc{different} for skill.
Skill templates have per-source quotas of 700/100/250 (S1, ordinal form), 500/75/200 (S1, rubric form) and
700/75/230 (S2) items; the two skill sources never share a quota, and a shortfall in one is not filled from the
other.
Any (template, question subject) key, i.e.\ the structure, instrument or skill criterion named in the stem, with at
least 20 candidates of which over 85\,\% fall into one answer cell is dropped, so the question alone does not predict
the answer. Options are shuffled so that the correct letter is exactly uniform over a--d within every split and
template.

\subsubsection{Audit}\label{sec:audit}

Table~\ref{tab:audit} lists representative errors that the clinician audit rounds found in
caption-lane items, and how each was resolved.
Every fix was kept as a hard check, so the released build contains no item that the check flags.

\begin{table}[!ht]
\caption{Representative errors found during the audit rounds and how each was fixed. Rd.: audit round in which the error was found. A hard check re-runs every fix on each build, and the build fails if any
item is flagged.}\label{tab:audit}
\footnotesize
\begin{tabular}{@{}lL{2.7cm}L{4.7cm}L{3.9cm}@{}}
\toprule
Rd. & Error & Example & Fix (hard check) \\
\midrule
1 & Object as actor &
  A specimen bag or a trocar counted as the instrument engaging tissue (T3, T4). &
  Object nouns never act; they are excluded as instruments. \\
1 & Release read as engagement &
  ``The grasper releases the gallbladder'' marked the grasper as engaged with it (T4). &
  A tool that the caption says is releasing the target is not engaged. \\
\midrule
2 & True distractor &
  A distractor interaction or goal that one of the two clips actually shows (C1, C3). &
  No distractor may be true of either video. \\
2 & Overlapping values as \textsc{different} &
  \emph{left} vs.\ \emph{upper left} keyed as a different position or entry side (A3, T2). &
  Overlapping values are never \textsc{different}, nor used as a distractor. \\
2 & Value from a negation &
  ``no unified tubular structure'' turned into a \emph{tubular} cystic duct (shape). &
  Values in negated clauses are dropped. \\
\midrule
3 & Absence from silence &
  ``Is the cystic duct tubular in both videos?'' answered ``only in Video~1'' because the other caption does not
  mention the shape. &
  Template disabled: shape does not change between clips. \\
\midrule
4 & Non-tissue interaction &
  ``holds the thread'' compared with ``sutures the small bowel'' as an interaction with tissue (C1). &
  Interaction options must act on tissue. \\
4 & Same manipulation as \textsc{different} &
  ``grasps the gallbladder'' vs.\ ``retracts the gallbladder'' keyed as different interactions (C1). &
  Interactions are compared by kind of action; \textsc{different} needs different kinds. \\
\midrule
5 & Reach read as engagement &
  ``(The hook is reaching the gallbladder)'' counted as the hook engaging the gallbladder (T3, T4). &
  Engagement requires a visible action on the target other than reaching. \\
5 & Same tissue, two names &
  Yellow lumpy fat named \emph{omentum} in one caption and \emph{mesentery} in the other, keyed as present in only
  one video (A1). &
  Omentum, mesentery and fat form one structure family; absence must hold for the whole family. \\
5 & Synonym distractor &
  \emph{lumpy} vs.\ \emph{lobulated}, which a viewer cannot tell apart (texture). &
  Template disabled (texture). \\
5 & Answer given by anatomy &
  ``Is the omentum fatty in both videos?'' is always \emph{yes}. &
  State \emph{fatty} removed. \\
5 & Answer given by the question &
  ``Is the triangular retractor engaged with the liver in both videos?'' is \emph{both} in 61 of 64 candidates
  (MultiBypass140). &
  Keys with over 85\,\% of candidates in one answer cell are dropped. \\
\bottomrule
\end{tabular}
\end{table}

\section{EndoDiffVQA statistics and examples}\label{sec:stats}

\subsection{Statistics}\label{sec:stats-numbers}

\begin{figure}[!ht]
\centering
\includegraphics[width=\textwidth]{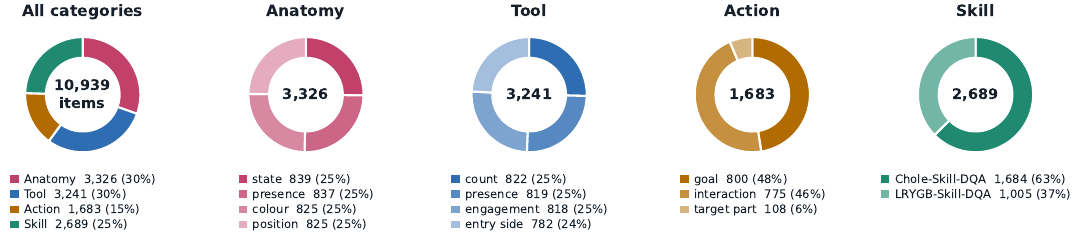}
\caption{Items per category and attribute in EndoDiffVQA.}\label{fig:stats}
\end{figure}

The release contains 10{,}939 items over 1{,}762 videos (7{,}289/1{,}350/2{,}300 train/val/test).
Fig.~\ref{fig:stats} shows the number of items per category and attribute, and Table~\ref{tab:duration} the clip
duration by annotation type.

\begin{table}[!ht]
\caption{Clip duration in EndoDiffVQA by annotation type, pooled over both procedures. Mean, variance and median
are in seconds; variance is the population variance.}\label{tab:duration}
\footnotesize
\begin{tabular}{@{}lrrrr@{}}
\toprule
Annotation type & Videos & Mean (s) & Variance (s$^2$) & Median (s) \\
\midrule
Captions (CholecT50, MultiBypass140)   & 1{,}442 &   9.2 &   1{,}224.8 &   2 \\
Skill scores (CholeScore, MultiBypass140) &   320 & 977.2 & 741{,}469.5 & 784 \\
\bottomrule
\end{tabular}
\end{table}

\subsection{Example items per template}\label{sec:examples}

\begin{figure}[!htbp]
\centering
\includegraphics[width=\textwidth]{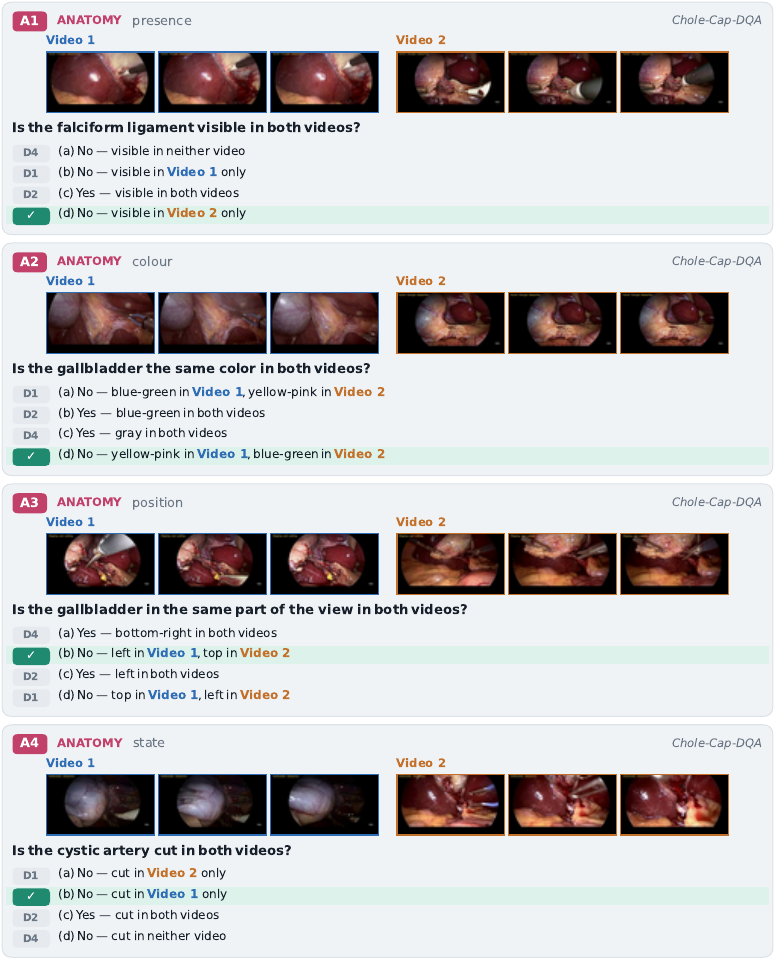}
\caption{Example items of the Anatomy templates A1 (presence), A2 (colour), A3 (position) and A4 (state). Three
frames per clip; the correct option is highlighted and each distractor is tagged with its error type.}
\label{fig:ex-anatomy}
\end{figure}

\begin{figure}[!htbp]
\centering
\includegraphics[width=\textwidth]{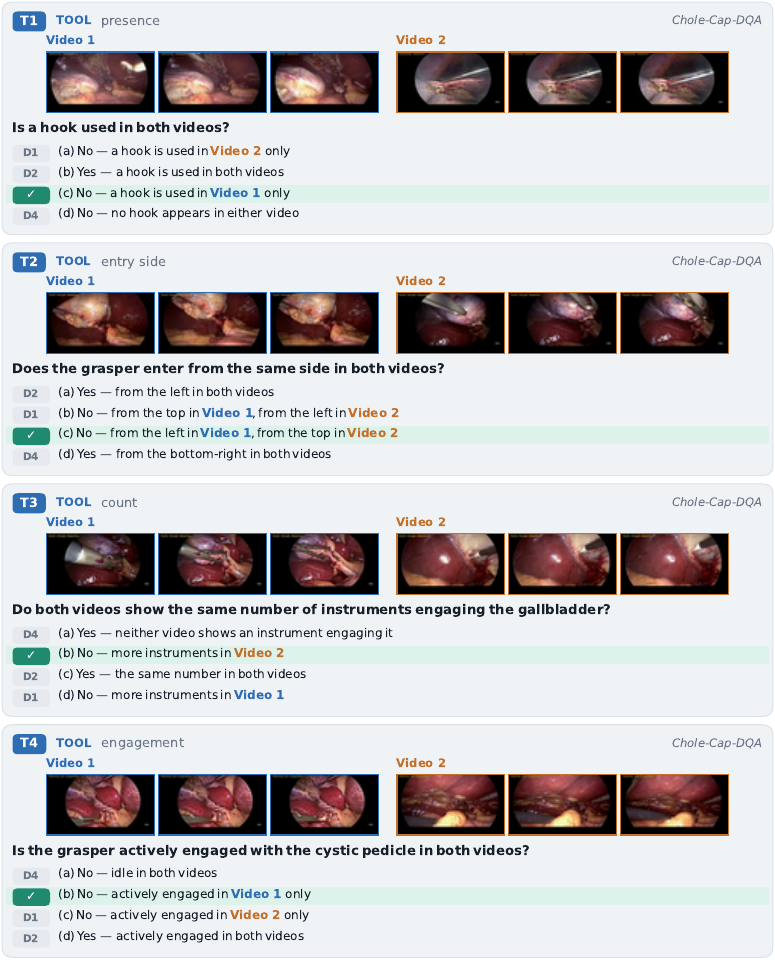}
\caption{Example items of the Tool templates T1 (presence), T2 (entry side), T3 (count) and T4 (engagement).}
\label{fig:ex-tool}
\end{figure}

\begin{figure}[!htbp]
\centering
\includegraphics[width=\textwidth]{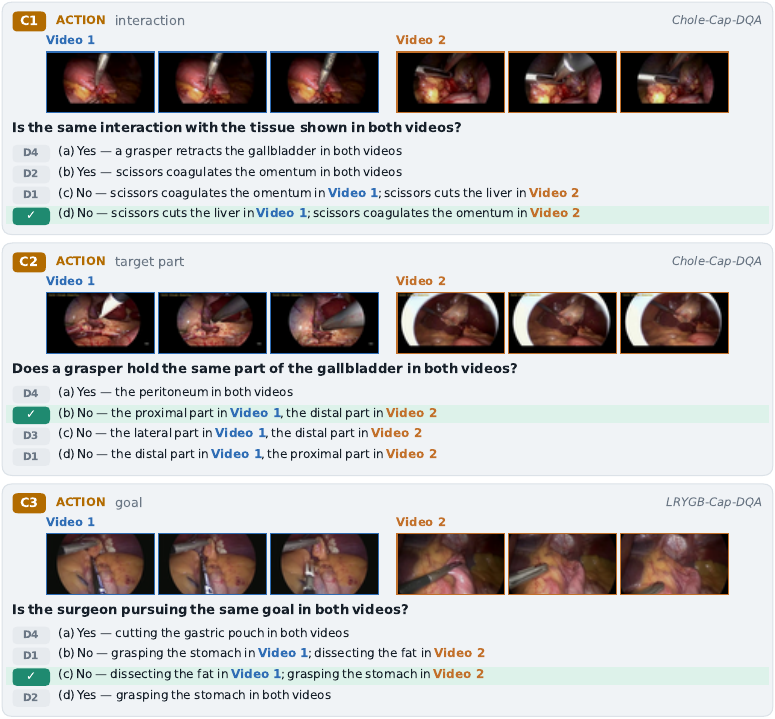}
\caption{Example items of the Action templates C1 (interaction), C2 (target part) and C3 (goal).}
\label{fig:ex-action}
\end{figure}

\begin{figure}[!htbp]
\centering
\includegraphics[width=\textwidth]{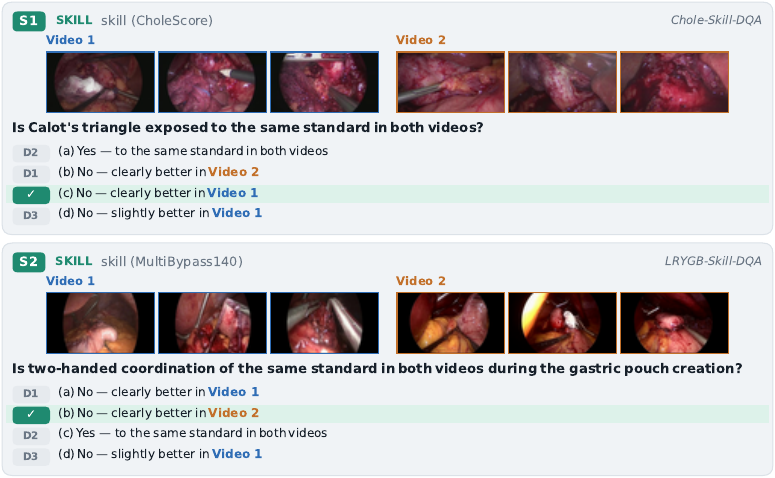}
\caption{Example items of the Skill templates S1 (\choleskill, CholeScore criteria) and S2 (\lrygbskill, GOALS
domains). The clips are phase-level and last minutes, so the three frames span a whole surgical step.}
\label{fig:ex-skill}
\end{figure}

Figs.~\ref{fig:ex-anatomy}--\ref{fig:ex-skill} show one released item for each of the 13 templates of
Table~\ref{tab:templates}, grouped by category. Each card shows three frames per clip, sampled at 15\,\%, 50\,\% and
85\,\% of the clip, the question stem and the four options, with the correct option highlighted and each
distractor tagged with its error type (D1 swap, D2 collapse, D3 substitute, D4 absent). The \emph{Video~1} /
\emph{Video~2} colours in the options match the frame borders. The caption-lane items are drawn from
\cholecap and \lrygbcap, the skill items from \choleskill and \lrygbskill; the subset is named at the top right of
each card.

\FloatBarrier

\section{Experiments and results}\label{sec:experiments}

\subsection{Baseline specifications}\label{sec:baselines}

This subsection specifies the four groups of compared systems (main paper, Section~3).

\subsubsection{Zero-shot vision-language models}

Open models of two sizes, Qwen3.5-9B and Qwen3.8-27B,
InternVL3.5-8B and InternVL3.5-30B, and Qwen2-VL-7B, plus two API models, GPT-5.5 and Gemini~3.6 Flash. We also run SurgVidLM as a surgical zero-shot model. 

\subsubsection{Fine-tuned vision-language models}

Qwen3.8-27B, InternVL3.5-8B, Qwen2-VL-7B and SurgVidLM are
fine-tuned with the common recipe: LoRA (rank 32, $\alpha=64$) on the language model with the vision encoder frozen,
learning rate $10^{-4}$ with cosine decay and 3\,\% warm-up, one epoch on train\,+\,validation at an effective batch
of 16, and the option letter as the target. The same recipe on our backbone, Qwen3.5-9B, is the \emph{letter-target}
baseline; adding one shared (Siamese) LoRA on the vision encoder and the merger gives the \emph{Siamese} baseline,
the direct counterpart of the pseudo-Siamese adaptation of Section~2.2.2 of the main paper.

\subsubsection{Surgical video-text encoders with a trained head}

PeskaVLP, and SurgCLIP are CLIP-style
encoders pre-trained on surgical video and lecture text. They cannot answer a comparative question on their own,
so we keep each encoder frozen, embed the two videos, the question stem and the four option texts, and train a
small option-selection head (1--3\,M parameters, identical for every encoder) that scores each option from
order-aware pair features $[v_1,v_2,v_1-v_2,v_1\odot v_2]$ together with the option and question embeddings. The head's
epochs, learning rate and dropout are chosen once on the validation split and reused for every encoder; the chosen
setting is then refit on train\,+\,validation with three seeds, and we report the mean.

\subsubsection{Surgical VQA architecture}

SurgViVQA (VideoMAE-B video encoder, BLIP text encoder, GPT-2 with LoRA) is
re-trained on EndoDiffVQA with a two-video adapter, at its native 8 frames per video and $224^2$ resolution, since
its released checkpoints answer free-text questions on other datasets.

\subsection{Qualitative results}\label{sec:qualitative}

\begin{figure}[p]
\centering
\includegraphics[width=\textwidth]{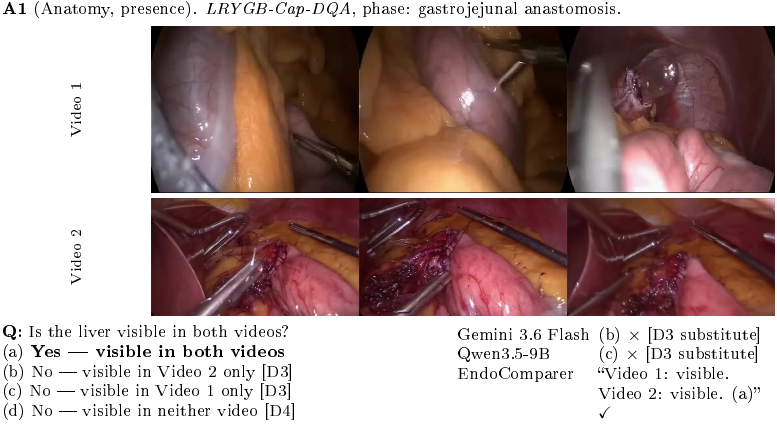}\par
\vspace{6pt}
\includegraphics[width=\textwidth]{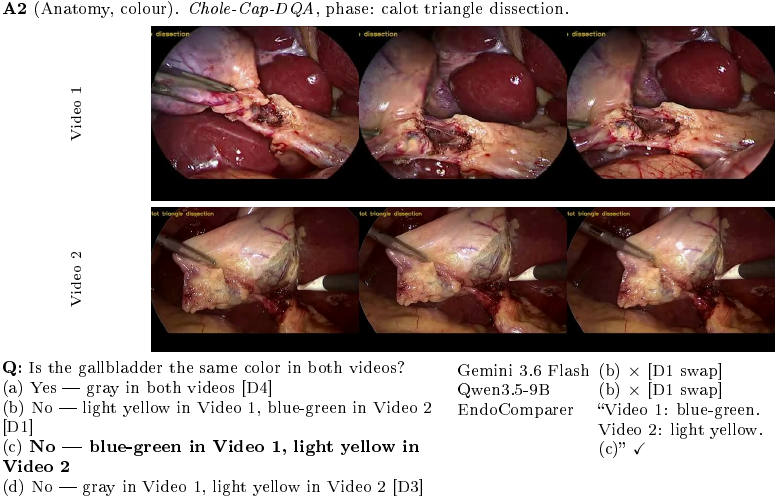}\par
\caption{Qualitative examples, templates A1 (Anatomy, presence) and A2 (Anatomy, colour) (legend in Section~\ref{sec:qualitative}).}\label{fig:qual-1}
\end{figure}

\begin{figure}[p]
\centering
\includegraphics[width=\textwidth]{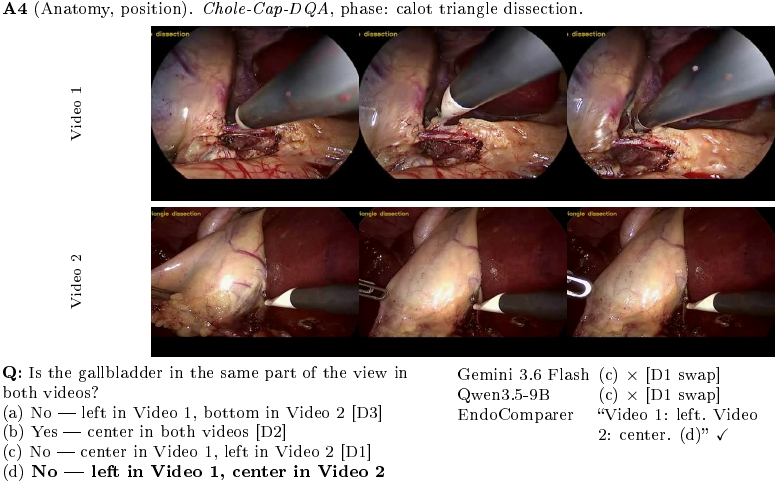}\par
\vspace{6pt}
\includegraphics[width=\textwidth]{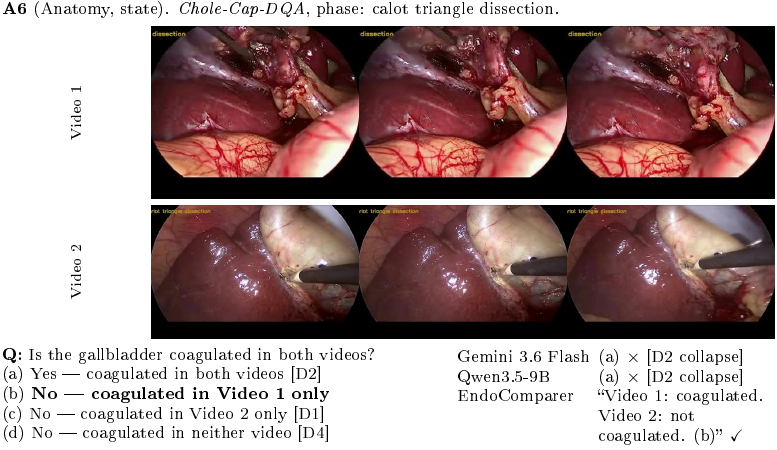}\par
\caption{Qualitative examples, templates A4 (Anatomy, position) and A6 (Anatomy, state) (legend in Section~\ref{sec:qualitative}).}\label{fig:qual-2}
\end{figure}

\begin{figure}[p]
\centering
\includegraphics[width=\textwidth]{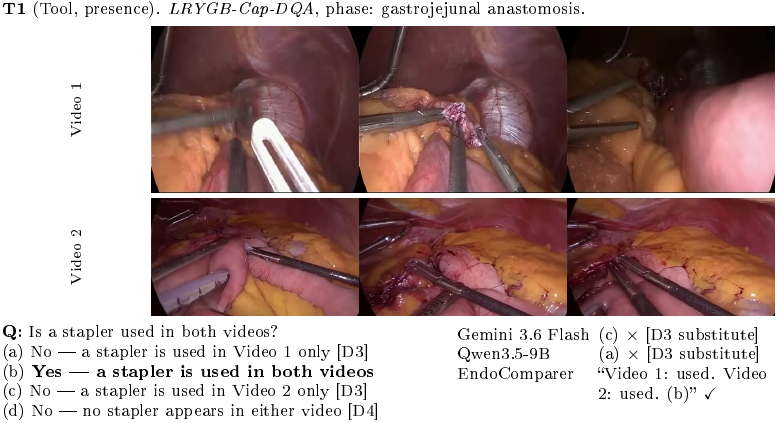}\par
\vspace{6pt}
\includegraphics[width=\textwidth]{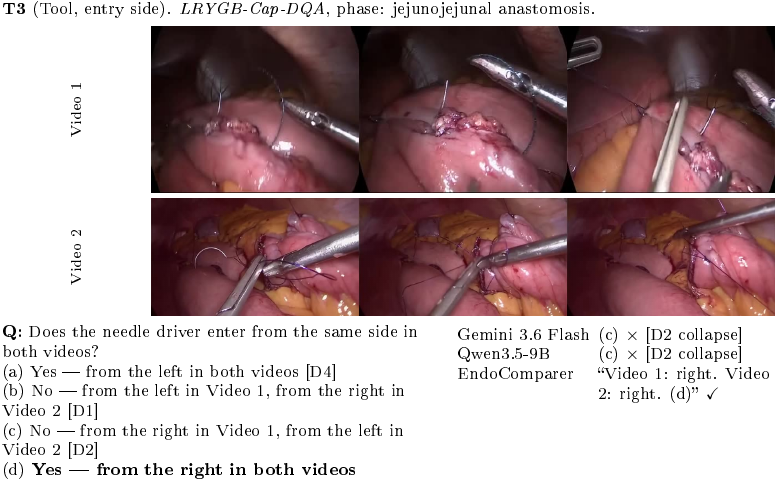}\par
\caption{Qualitative examples, templates T1 (Tool, presence) and T3 (Tool, entry side) (legend in Section~\ref{sec:qualitative}).}\label{fig:qual-3}
\end{figure}

\begin{figure}[p]
\centering
\includegraphics[width=\textwidth]{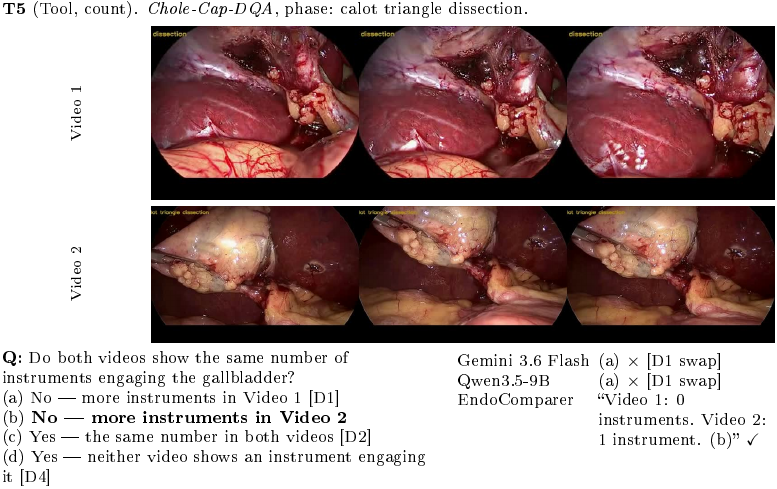}\par
\vspace{6pt}
\includegraphics[width=\textwidth]{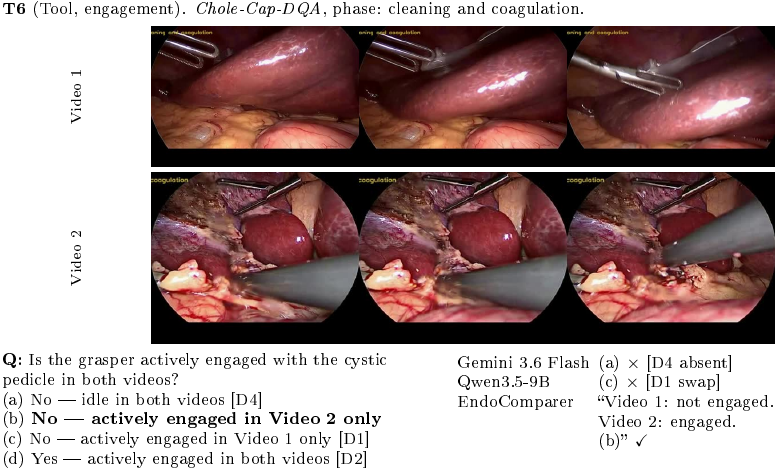}\par
\caption{Qualitative examples, templates T5 (Tool, count) and T6 (Tool, engagement) (legend in Section~\ref{sec:qualitative}).}\label{fig:qual-4}
\end{figure}

\begin{figure}[p]
\centering
\includegraphics[width=\textwidth]{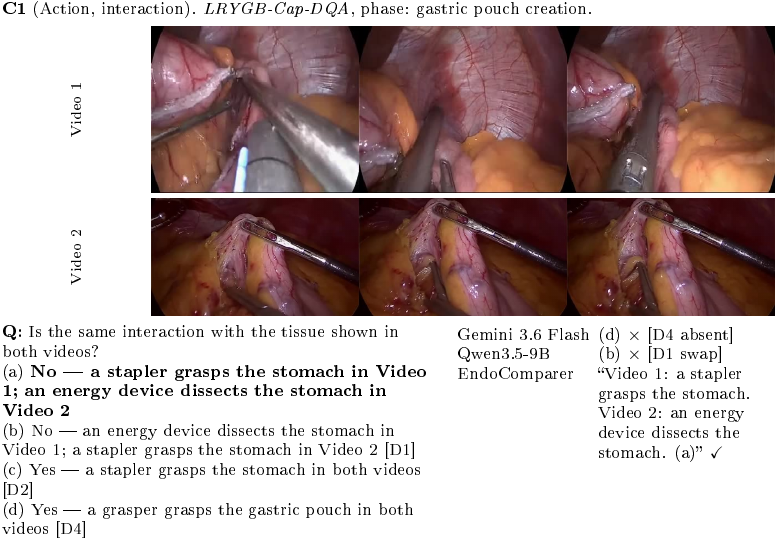}\par
\vspace{6pt}
\includegraphics[width=\textwidth]{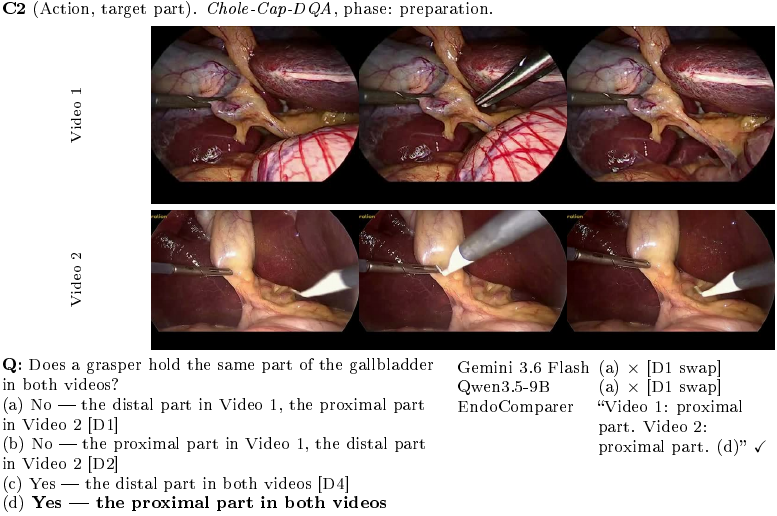}\par
\caption{Qualitative examples, templates C1 (Action, interaction) and C2 (Action, target part) (legend in Section~\ref{sec:qualitative}).}\label{fig:qual-5}
\end{figure}

\begin{figure}[p]
\centering
\includegraphics[width=\textwidth]{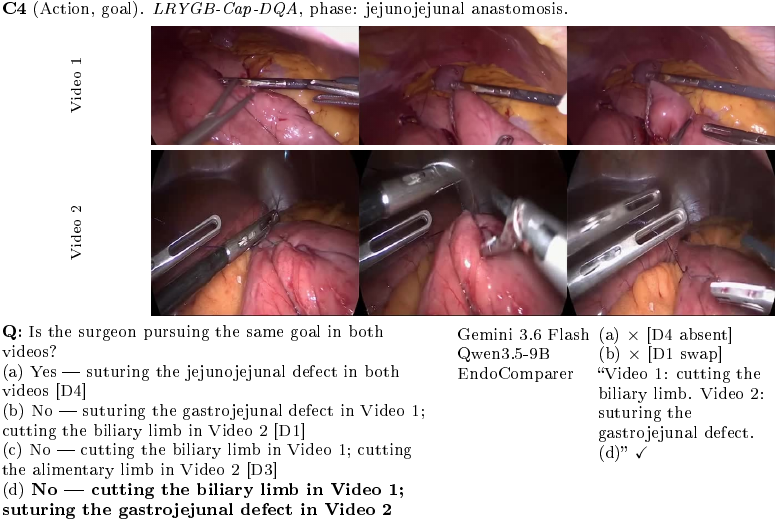}\par
\vspace{6pt}
\includegraphics[width=\textwidth]{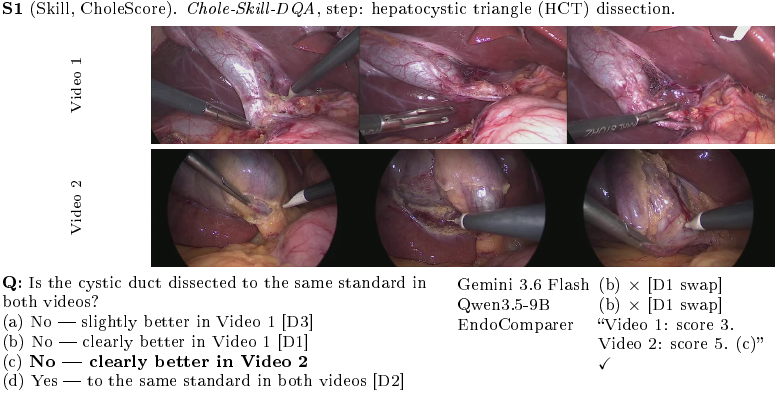}\par
\caption{Qualitative examples, templates C4 (Action, goal) and S1 (Skill, CholeScore) (legend in Section~\ref{sec:qualitative}).}\label{fig:qual-6}
\end{figure}

\begin{figure}[p]
\centering
\includegraphics[width=\textwidth]{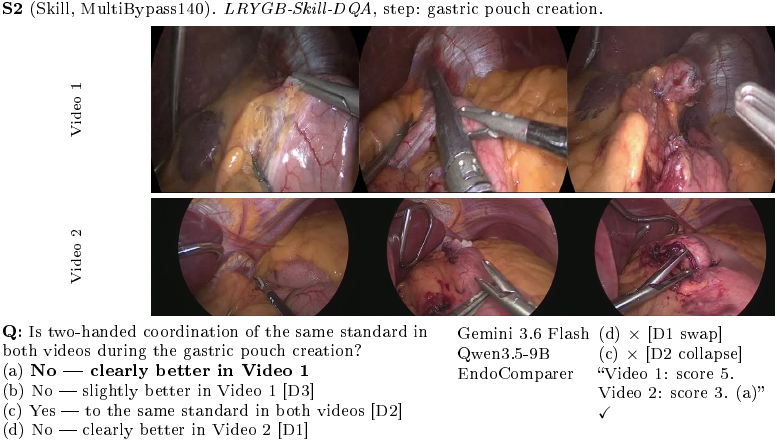}\par
\caption{Qualitative example, template S2 (Skill, MultiBypass140) (legend in Section~\ref{sec:qualitative}).}\label{fig:qual-7}
\end{figure}

Figures~\ref{fig:qual-1}--\ref{fig:qual-7} show one test item per question template, with three frames
per video and the outputs of the strongest zero-shot model (Gemini~3.6 Flash), the finetuned backbone (Qwen3.5-9B)
and EndoComparer. Each panel shows three frames per video (start, middle and end of the clip; for A1 the frames were chosen
to show the queried structure, which the uniformly spaced frames miss), the question with its four options,
the gold option in bold and each distractor tagged with its type (D1 swap, D2 collapse, D3 substitute, D4 absent),
and the three outputs. The baselines output only an option letter; for EndoComparer we reproduce the full response,
i.e.\ its Perceive-then-Compare chain of thought followed by the chosen letter. The most frequent baseline error is the \textbf{D1 swap}: the two facts are perceived but
attached to the wrong videos (A2, C2, T5, S1, S2), which is exactly the video-confusion failure that the pseudo-Siamese
adapter and the per-video statements of the chain of thought address. The \textbf{D3 substitute} on A1 is of the same
kind: the liver is visible in both clips, yet each baseline attributes it to a single video, and not the same one. The
\textbf{D2 collapse} (A4, T1, T3) and \textbf{D4 absent} (A6, C1, C4, T6) errors show the baselines either defaulting
to ``same'' or describing content that appears in neither video.

\end{document}